\documentclass[letterpaper]{article} 
\usepackage[preprint]{aaai2027}    

\usepackage[hyphens]{url}            
\usepackage{graphicx}                
\usepackage{natbib}                  
\usepackage{caption}                 
\usepackage{amsmath,amssymb,amsfonts}

\usepackage{algorithm}
\usepackage{algorithmic}

\usepackage[caption=false,font=footnotesize]{subfig}
\usepackage{booktabs}
\usepackage{multirow}
\usepackage{array}
\usepackage{xcolor}

\title{LIRA: Local Cross-Layer Information Routing for Vision-Language-Action Decoding}
\author{
    Zhewei Zhang\textsuperscript{*},
    Puyue Wang\textsuperscript{*},
    Guanren Qiao\textsuperscript{*},
    Yijie Weng,
    Jiawei Hu,
    Guo Li,
    Lujia Wang,
    Junyan Wang,
    Tao Gu,
    Hongliang Lu\textsuperscript{\textdagger},
    Guiliang Liu,
    Hong Jia,
    Xinhu Zheng\textsuperscript{\textdaggerdbl}
}

\affiliations{
    \textsuperscript{*}Equal Contribution,\quad
    \textsuperscript{\textdagger}Project Lead,\quad
    \textsuperscript{\textdaggerdbl}Corresponding Author
}

\begin{document}

\maketitle

\begin{abstract}
Vision-Language-Action (VLA) models transform representations from pretrained vision-language models (VLMs) into robot actions, yet the interface that routes intermediate VLM features into action decoders remains underexplored. Existing designs either expose only a narrow part of the representation hierarchy or rigidly match each decoder block to one VLM layer, restricting access to complementary task evidence across depths. We introduce \textsc{LIRA}, a local cross-layer action-conditioning mechanism that formulates VLM-to-action conditioning as depth-aware information routing. \textsc{LIRA} operates on task-token features and LIRA Query features derived from intermediate VLM states, then assigns each Parallel Fusion Block a depth-aligned local window centered on its corresponding VLM layer. Parallel Fusion Blocks aggregate neighboring LIRA Query features and integrate them with task-token features and proprioceptive inputs before action prediction. This routing interface leaves the backbone architecture, action decoder, and supervised training recipe unchanged. Across LIBERO, LIBERO-Plus, CALVIN ABC$\rightarrow$D, and real-world manipulation, \textsc{LIRA} improves the principal aggregate metrics over the VLA-Adapter baseline under the same 0.5B-parameter configuration. In zero-shot transfer to LIBERO-Plus, \textsc{LIRA} increases average success from 59.1\% to 78.0\%, an 18.9-point gain indicating improved robustness under controlled distribution shifts. 
\end{abstract}

\section{Introduction}
\label{sec:introduction}

Vision-Language-Action (VLA) models map visual observations and
language instructions to robot actions by reusing pretrained
vision-language model (VLM) representations
~\citep{brohan2023rt1,zitkovich2023rt2,ghosh2024octo,
kim2024openvla,fang2026molmoact2}.
Although larger robot datasets and generalist policies have advanced
this paradigm~\citep{openx2023rtx,khazatsky2024droid}, the
VLM-to-action interface that routes intermediate VLM representations
into the action decoder remains underexplored.
This interface determines which visual, linguistic, and task evidence
is available during action generation
~\citep{gao2026vla,kong2026affordvla,
huang2026thinkact,denggraspvla}.

\begin{figure}[!t]
    \centering
    \includegraphics[width=\columnwidth]{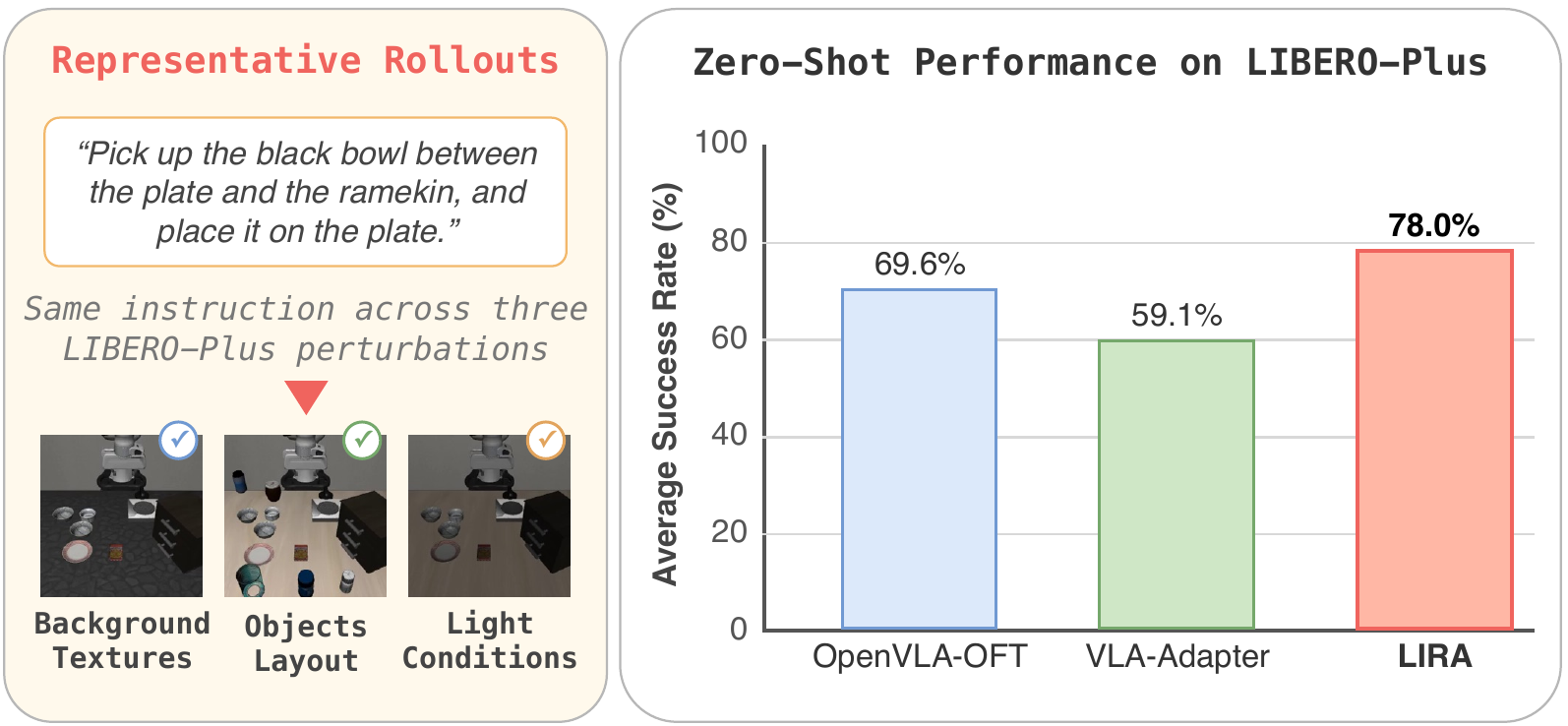}
    \caption{
Zero-shot transfer to LIBERO-Plus.
Left: representative \textsc{LIRA} rollouts for the same language
instruction under Background, Layout, and Light perturbations; all three
are completed successfully without LIBERO-Plus fine-tuning.
Right: overall task-level success over all 10,030 LIBERO-Plus tasks.
\textsc{LIRA} achieves 78.0\%, outperforming OpenVLA-OFT and VLA-Adapter
by 8.4 and 18.9 percentage points, respectively.
}
    \label{fig:teaser}
\end{figure}

\begin{table*}[t]
\centering

{\small
\setlength{\tabcolsep}{6pt}
\begin{tabular*}{\textwidth}{
    @{\extracolsep{\fill}}lcccc@{}
}
\toprule
Method
& Action module
& Multi-layer VLM
& Depth-aligned
& Local cross-layer \\
\midrule

RT-2~\citep{zitkovich2023rt2}
& $\times$ & $\times$ & $\times$ & $\times$ \\

OpenVLA~\citep{kim2024openvla}
& $\times$ & $\times$ & $\times$ & $\times$ \\

OpenVLA-OFT~\citep{kim2025fine}
& $\checkmark$ & $\times$ & $\times$ & $\times$ \\

GR00T N1~\citep{bjorck2025gr00t}
& $\checkmark$ & $\times$ & $\times$ & $\times$ \\

$\pi_{0.5}$~\citep{black2025pi05}
& $\checkmark$ & $\checkmark$ & $\checkmark$ & $\times$ \\

VLA-Adapter~\citep{wang2026vlaadapter}
& $\checkmark$ & $\checkmark$ & $\checkmark$ & $\times$ \\

\textbf{\textsc{LIRA} (Ours)}
& $\checkmark$ & $\checkmark$ & $\checkmark$ & $\checkmark$ \\

\bottomrule
\end{tabular*}
}

\caption{
Comparison of representative VLM-to-action interfaces.
``Action module'' denotes an action-specific head, expert, or policy;
``Multi-layer VLM'' denotes conditioning on representations from multiple
VLM depths; ``Depth-aligned'' denotes associating each decoder block with
a corresponding VLM depth; and ``Local cross-layer'' denotes expanding
the conditioning source to adjacent VLM layers.
}
\label{tab:interface_taxonomy}
\end{table*}

Table~\ref{tab:interface_taxonomy} summarizes representative
VLM-to-action interfaces.
Existing systems either use final-layer features, apply one-to-one
layer-aligned conditioning, or aggregate representations from multiple
VLM layers
~\citep{kim2025fine,pertsch2025fast,bjorck2025gr00t,
black2024pi_0,li2026rotvla,cen2025worldvla,song2025pd,
wang2026blockvla,zhao2025cot}.
Final-layer conditioning exposes only terminal representations;
one-to-one layer-aligned conditioning restricts each decoder block to
one matched VLM layer; and global aggregation does not explicitly
preserve local structure across VLM depth.
These restrictions may be particularly limiting under controlled
distribution shifts.
LIBERO-Plus systematically varies camera viewpoints, robot
initialization, language instructions, lighting, backgrounds, sensor
noise, and object layouts~\citep{fei2026liberoplus}, providing a
controlled evaluation of robustness under such shifts.

These limitations motivate a central question:
\textit{How should intermediate VLM representations be routed into the
action decoder?}
Our hypothesis is that adjacent VLM layers provide complementary task
evidence for action prediction at a given decoder depth.
We therefore propose \textsc{LIRA}, a local cross-layer routing
mechanism built on VLA-Adapter~\citep{wang2026vlaadapter}.
\textsc{LIRA} appends learnable LIRA Query tokens to the VLM and
extracts their layerwise features.
For each Parallel Fusion Block (PFB), it aggregates LIRA Query features
from a centered window of adjacent VLM layers, while task-token features
remain aligned with the corresponding VLM depth.
This design gives each PFB access to neighboring representations
without discarding the depth alignment of the original interface.

We evaluate \textsc{LIRA} on LIBERO, LIBERO-Plus, CALVIN
ABC$\rightarrow$D, and real-world manipulation tasks
~\citep{liu2023libero,fei2026liberoplus,mees2022calvin}.
Under matched model settings, \textsc{LIRA} improves the
VLA-Adapter baseline across all three simulation benchmarks.
The largest gain occurs in zero-shot transfer to LIBERO-Plus, where
\textsc{LIRA} improves overall task-level success by 18.9 percentage
points and exceeds OpenVLA-OFT by 8.4 points
(Figure~\ref{fig:teaser}).
Routing-topology, window-width, and query-budget ablations further
support the proposed depth-locality prior, while the physical-robot
evaluation extends the comparison beyond simulation.

\textbf{Our contributions are as follows:}
\begin{enumerate}
    \renewcommand{\labelenumi}{\arabic{enumi})}

    \item We recast VLM-to-action conditioning as a depth-aware
    information-routing problem and introduce a locality prior over VLM
    depth: each PFB should access a centered local window of adjacent
    VLM layers rather than only one matched layer or the full VLM
    hierarchy.

    \item Guided by this prior, we propose \textsc{LIRA}, a local
    cross-layer action-conditioning mechanism that routes LIRA Query
    features from a depth-aligned local window to each PFB while retaining
   the layer-aligned task-token branch.
    This design provides structured local context across VLM depth and
    introduces no additional trainable parameters relative to VLA-Adapter.

    \item Under a matched 0.5B-backbone configuration, \textsc{LIRA}
    improves the VLA-Adapter baseline across LIBERO, CALVIN
    ABC$\rightarrow$D, and LIBERO-Plus.
    In zero-shot transfer to LIBERO-Plus, it raises overall task-level
    success from 59.1\% to 78.0\%, while routing-topology,
    routing-window-width, and query-budget ablations further support
    the proposed locality prior.
\end{enumerate}
\begin{figure*}[!t]
    \centering
    \includegraphics[width=\textwidth]{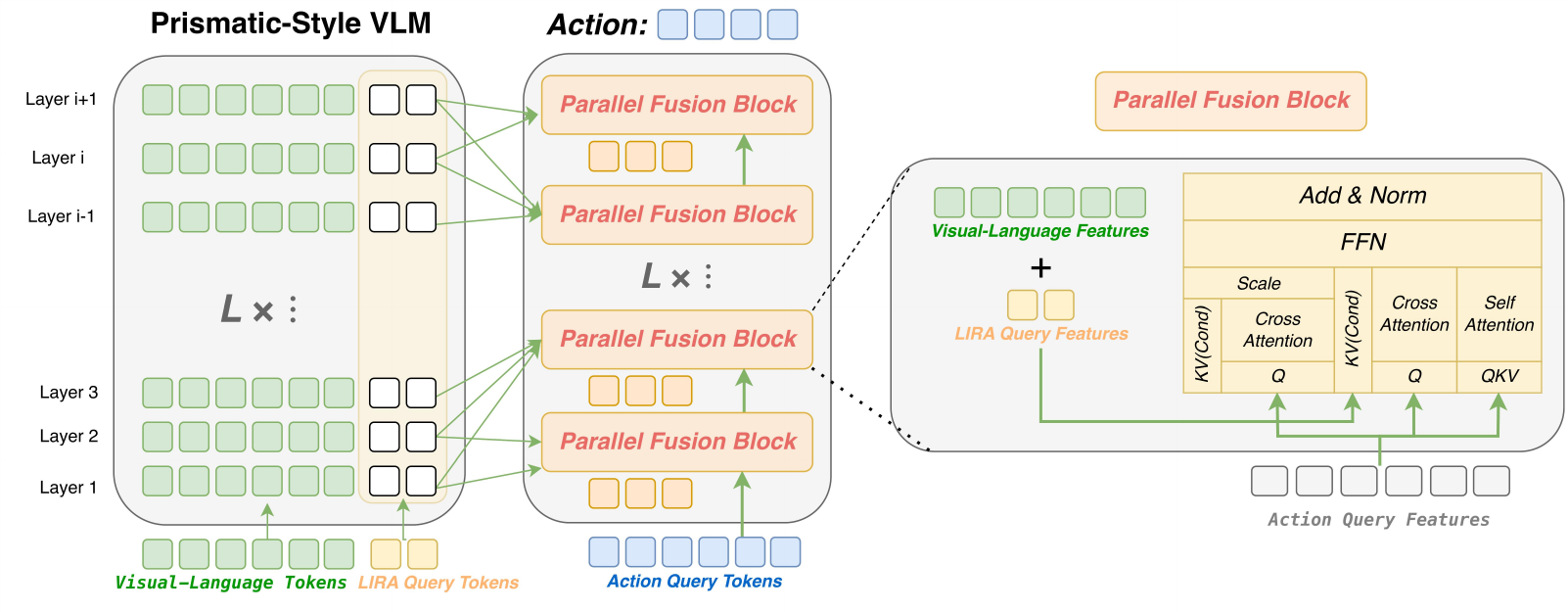}
     \caption{
    Overview of \textsc{LIRA}.
    The Prismatic-style VLM processes visual-language task tokens together
    with learnable LIRA Query tokens, while Action Query tokens initialize
    an action decoder composed of $L$ Parallel Fusion Blocks (PFBs).
    Task-token features remain aligned with the corresponding VLM depth,
    while each PFB receives LIRA Query features aggregated from a centered
    local window of adjacent VLM layers.
    Each PFB uses Action Query features to fuse the two conditioning sources
    before action prediction.
    The proprioceptive branch is omitted for clarity.
    }
    \label{fig:framework}
\end{figure*}

\section{Related Work}
\label{sec:related}

\paragraph{Vision-Language-Action Models for Robot Manipulation.}
Early generalist robot policies and embodied multimodal models,
including Gato, RT-1, PaLM-E, and Octo, demonstrated the potential of
scalable sequence modeling and multimodal pretraining for robotic
control
~\citep{reed2022gato,brohan2023rt1,driess2023palme,ghosh2024octo}.
VLM-based VLA models such as RT-2 and OpenVLA further represent robot
actions as tokens and generate them using language-model decoders
~\citep{zitkovich2023rt2,kim2024openvla}.
Recent work improves action generation through parallel continuous
decoding, flow-matching experts, reinforcement-learning fine-tuning,
compressed action tokenization, and compact action modules
~\citep{kim2025fine,black2025pi05,pertsch2025fast,bjorck2025gr00t,
hung2025nora,shukor2025smolvla,cui2025openhelix,
wang2026vlaadapter,lu2025vla,li2025simplevla,
qiao2026focusthencontact}.
These directions primarily modify action representations, training
recipes, model scale, or action-generation architectures.
Our work instead studies VLM-to-action conditioning: how intermediate
VLM representations are routed into the action decoder.

\paragraph{VLM-to-Action Interfaces and Intermediate Representations.}
VLA architectures expose pretrained VLM representations to action
modules through different interfaces.
OpenVLA-OFT maps final-layer VLM representations to continuous actions
using an MLP action head~\citep{kim2025fine}, whereas GR00T N1
conditions its diffusion transformer on features extracted from a
single intermediate VLM layer~\citep{bjorck2025gr00t}.
In contrast, the action expert in $\pi_{0.5}$ interacts with VLM
representations throughout the Transformer depth
~\citep{black2025pi05}.
VLA-Adapter explicitly exposes multi-layer VLM representations and
conditions each Parallel Fusion Block (PFB) on task-token and VLM-side
query features from one matched VLM layer
~\citep{wang2026vlaadapter}.

VLA-Adapter is the closest architectural baseline because it implements
one-to-one layer-aligned conditioning between VLM depth and PFB depth.
\textsc{LIRA} replaces its single-layer query conditioning with a
centered local window while retaining the remaining architecture and
training objective.

\paragraph{Cross-Layer Aggregation.}
More broadly, cross-layer aggregation has been studied as a means of
reusing information distributed across network depth.
Deep Layer Aggregation hierarchically merges intermediate
representations, whereas sparse aggregation emphasizes selecting a
limited subset of preceding layers rather than densely combining all features~\citep{yu2018deep,zhu2018sparsely}.
These works motivate structured feature reuse;
\textsc{LIRA} instead routes intermediate VLM representations
to a depth-aligned action decoder.

\section{Methodology}
\label{sec:method}

\textsc{LIRA} formulates VLM-to-action conditioning as a depth-aware
information-routing problem.
Existing layer-aligned interfaces expose multiple VLM depths to the
action decoder, but restrict each decoder block to the features of one
matched VLM layer.
Our key idea is to condition each Parallel Fusion Block (PFB) on a local
neighborhood of adjacent VLM layers.
This provides complementary representations across nearby depths while
preserving the local organization of the VLM hierarchy.

Figure~\ref{fig:framework} illustrates the overall architecture.
We first introduce the depth-aligned VLM-to-action interface and then
describe the proposed local cross-layer routing mechanism.

\subsection{Depth-Aligned VLM-to-Action Interface}
\label{subsec:vlm_to_action_interface}

At control step $t$, the agent receives a third-person image $X_t^v$, a
gripper-view image $X_t^g$, a language instruction $u_t$, and the robot
proprioceptive state $P_t$.
The images and instruction are encoded as visual-language tokens.
Following VLA-Adapter~\citep{wang2026vlaadapter}, the VLM input also
contains $M$ learnable VLM-side query tokens, which we refer to as
\emph{LIRA Query tokens} throughout this paper.

Let the VLM contain $L$ Transformer layers.
From VLM layer $\ell$, we extract task-token features
$T_t^{(\ell)}$ and LIRA Query features
$R_t^{(\ell)}\in\mathbb{R}^{M\times d_v}$, where $d_v$ is the VLM
hidden dimension.
The task-token features contain the visual-language representations,
whereas the LIRA Query features provide a conditioning
interface between the VLM and the action decoder.
The proprioceptive state is encoded by a learnable projection
$\phi_p$, producing the embedding $E_t^p=\phi_p(P_t)$.

The action decoder contains $L$ PFBs and is initialized by a sequence
of Action Query tokens.
We use $X_t^{(0)}$ to denote their initial embeddings and
$X_t^{(i)}$ to denote the Action Query features produced after the
$i$-th PFB.
Within each PFB, Bridge Attention updates the Action Query features
through self-attention and cross-attention to the VLM conditioning
features.
The task-token features provide visual-language context, while the LIRA
Query features, together with the proprioceptive embedding, provide
query-based action conditioning.

In the original one-to-one layer-aligned interface, PFB $i$ receives
both $T_t^{(i)}$ and $R_t^{(i)}$ from its matched VLM layer.
Although this design establishes a correspondence between VLM depth and
action-decoder depth, each PFB can access LIRA Query features from only
one layer.
Consequently, potentially complementary representations in adjacent
VLM layers are not available to the same PFB.

\begin{table*}[t]
\centering

{\small
\setlength{\tabcolsep}{6pt}
\begin{tabular*}{\textwidth}{
    @{\extracolsep{\fill}}lcccccc@{}
}
\toprule
Method & Params & Spatial & Object & Goal & Long & Avg. \\
\midrule

UnifiedVLA~\citep{wang2025unified}
& 8.5B & 95.4 & 98.8 & 93.6 & 94.0 & 95.5 \\

OpenVLA~\citep{kim2024openvla}
& 7B & 84.7 & 88.4 & 79.2 & 53.7 & 76.5 \\

UniVLA~\citep{bu2025univla}
& 7B & 96.5 & 96.8 & 95.6 & 92.0 & 95.2 \\

OpenVLA-OFT~\citep{kim2025fine}
& 7B & 97.6 & 98.4 & 97.9 & 94.5 & 97.1 \\

$\pi_0$~\citep{black2024pi_0}
& 3B & 96.8 & 98.8 & 95.8 & 85.2 & 94.2 \\

$\pi_0$-FAST~\citep{pertsch2025fast}
& 3B & 96.4 & 96.8 & 88.6 & 60.2 & 85.5 \\

AffordanceVLA (full)~\citep{yu2026affordancevla}
& 4B & 98.6 & 98.4 & 96.2 & 89.8 & 95.8 \\

VLA-Adapter~\citep{wang2026vlaadapter}
& 0.5B & 97.8 & 99.2 & 97.2 & 95.0 & 97.3 \\

VLA-Adapter-Pro~\citep{wang2026vlaadapter}
& 0.5B & \textbf{99.6} & 99.6 & 98.2 & 96.4 & 98.5 \\

\textbf{\textsc{LIRA} (Ours)}
& 0.5B
& \textbf{99.6}
& \textbf{99.8}
& \textbf{98.5}
& \textbf{97.6}
& \textbf{98.9} \\

\bottomrule
\end{tabular*}
}

\caption{
Comparison on LIBERO. Success rates are reported in \%.
Reported systems with different training pipelines provide benchmark
context; VLA-Adapter is the controlled architectural baseline for
\textsc{LIRA}.
}
\label{tab:libero_main}
\end{table*}

\begin{table*}[t]
\centering

{\small
\setlength{\tabcolsep}{6pt}
\begin{tabular*}{\textwidth}{
    @{\extracolsep{\fill}}lccccccc@{}
}
\toprule
Method & Params & 1 & 2 & 3 & 4 & 5 & Avg. len. \\
\midrule

UnifiedVLA~\citep{wang2025unified}
& 8.5B & 98.9 & 94.8 & 89.0 & 82.8 & 75.1 & 4.41 \\

OpenVLA~\citep{kim2024openvla}
& 7B & 91.3 & 77.8 & 62.0 & 52.1 & 43.5 & 3.27 \\

UniVLA~\citep{bu2025univla}
& 7B & 95.5 & 85.8 & 75.4 & 66.9 & 56.5 & 3.80 \\

OpenVLA-OFT~\citep{kim2025fine}
& 7B & 96.9 & 92.0 & 85.7 & 80.4 & 72.9 & 4.28 \\

AffordanceVLA (full)~\citep{yu2026affordancevla}
& 4B
& 96.8
& 92.0
& 87.5
& 80.8
& 75.9
& 4.33 \\

VLA-Adapter~\citep{wang2026vlaadapter}
& 0.5B & \textbf{99.1} & 94.6 & 88.8 & 82.8 & 76.5 & 4.42 \\

VLA-Adapter-Pro~\citep{wang2026vlaadapter}
& 0.5B & 98.5 & 95.0 & 90.5 & 85.3 & 80.0 & 4.50 \\

\textbf{\textsc{LIRA} (Ours)}
& 0.5B
& 98.8
& \textbf{95.3}
& \textbf{91.0}
& \textbf{85.8}
& \textbf{80.5}
& \textbf{4.52} \\

\bottomrule
\end{tabular*}
}

\caption{
Comparison on CALVIN ABC$\rightarrow$D. Success rates are reported in
\% for completing one to five tasks consecutively; Avg. len. denotes
the average completed sequence length.
}
\label{tab:calvin_main}
\end{table*}

\subsection{Local Cross-Layer Routing}
\label{subsec:local_cross_layer_routing}

Representations evolve progressively across the VLM hierarchy.
Adjacent layers may emphasize different aspects of the same input,
ranging from local visual details to object relations and
instruction-conditioned semantics.
Conditioning a PFB on only one layer can therefore omit useful nearby
information.
Conversely, aggregating representations from all VLM layers may mix
features from distant depths and obscure their local structure.

To balance these two extremes, \textsc{LIRA} introduces a
depth-aligned local routing window.
For PFB $i$, we select a neighborhood of VLM layers centered on its
matched layer and concatenate their LIRA Query features:
\begin{equation}
\begin{aligned}
\mathcal{W}_r(i)
&=
\left\{
\ell\in\{1,\ldots,L\}
\;\middle|\;
|\ell-i|\leq r
\right\},\\
\widetilde{R}_t^{(i)}
&=
\operatorname{Concat}_{\mathrm{tok}}
\left(
\left\{
R_t^{(\ell)}
\right\}_{\ell\in\mathcal{W}_r(i)}
\right).
\end{aligned}
\label{eq:lira_routing}
\end{equation}
The selected features are concatenated along the token dimension in
ascending layer order.
We use $r=1$ by default, corresponding to a centered three-layer window
for an interior PFB.
At the first and last layers, the window is clipped to the valid
boundary, resulting in a two-layer neighborhood.

The routed LIRA Query representation is used to condition the
corresponding PFB:
\begin{equation}
X_t^{(i)}
=
\operatorname{PFB}_{i}
\left(
X_t^{(i-1)};
T_t^{(i)},
\widetilde{R}_t^{(i)},
E_t^p
\right).
\label{eq:lira_conditioning}
\end{equation}
The matched task-token features $T_t^{(i)}$ provide a depth-aligned
semantic anchor, while $\widetilde{R}_t^{(i)}$ supplies complementary
information from the local VLM neighborhood.
Each PFB therefore receives a structured multi-depth representation
centered on its corresponding VLM layer.

For an interior PFB, the routed representation contains $3M$ LIRA Query
features; boundary PFBs receive $2M$ features.
Because the selected features are concatenated along the token
dimension, they can be consumed directly by the existing cross-attention
operations.
The routing mechanism therefore introduces no additional projection
layers or trainable parameters.

The routing-window width provides an explicit control over the amount
of cross-layer context.
A width of one recovers single-layer conditioning, the default width of
three covers the matched layer and its immediate neighbors, and larger
windows progressively approach global aggregation.
This design allows \textsc{LIRA} to incorporate complementary
representations across depth without flattening the entire VLM
hierarchy into a single global representation.

\subsection{Action Prediction and Training Objective}
\label{subsec:training_objective}

After the final PFB, an output head maps the resulting Action Query
features to a continuous action chunk of horizon $H$.
Let
$\hat{\mathbf{a}}_{t:t+H-1}
=\pi_\theta(X_t^v,X_t^g,u_t,P_t)$
denote the predicted action chunk, and let
$\mathbf{a}_{t:t+H-1}$ denote the corresponding ground-truth actions.
The model is trained using the supervised continuous-action objective
\begin{equation}
\mathcal{L}_{\mathrm{act}}
=
\mathbb{E}_{\mathcal{D}}
\left[
\frac{1}{H}
\sum_{h=0}^{H-1}
\left\|
\hat{\mathbf{a}}_{t+h}
-
\mathbf{a}_{t+h}
\right\|_1
\right],
\label{eq:training_objective}
\end{equation}
where $\mathcal{D}$ denotes the demonstration dataset.
The same objective and training recipe are used for the controlled
VLA-Adapter baseline and \textsc{LIRA}, allowing the experiments to
isolate the effect of local cross-layer routing.

\begin{table*}[t]
\centering

{\small
\setlength{\tabcolsep}{4pt}
\begin{tabular*}{\textwidth}{
    @{\extracolsep{\fill}}lc*{7}{c}c@{}
}
\toprule
Method
& Params
& Camera
& Robot
& Language
& Light
& Background
& Noise
& Layout
& Overall \\
\midrule

\multicolumn{10}{c}{\textit{Zero-shot transfer}} \\
\midrule

OpenVLA~\citep{kim2024openvla}
& 7B
& 0.8 & 3.5 & 23.0 & 8.1
& 34.8 & 15.2 & 28.5
& 15.6 \\

OpenVLA-OFT~\citep{kim2025fine}
& 7B
& 56.4 & 31.9 & 79.5 & 88.7
& 93.3 & 75.8 & 74.2
& 69.6 \\

UniVLA~\citep{bu2025univla}
& 7B
& 1.8 & 46.2 & 69.6 & 69.0
& 81.0 & 21.2 & 31.9
& 42.9 \\

$\pi_{0}$~\citep{black2024pi_0}
& 3B
& 13.8 & 6.0 & 58.8 & 85.0
& 81.4 & 79.0 & 68.8
& 53.6 \\

$\pi_{0}$-FAST~\citep{pertsch2025fast}
& 3B
& 65.1 & 21.6 & 61.0 & 73.2
& 73.2 & 74.4 & 68.8
& 61.6 \\

VLA-Adapter~\citep{wang2026vlaadapter}
& 0.5B
& 36.2 & 37.9 & 74.6 & 70.6
& 76.1 & 58.0 & 69.7
& 59.1 \\

\textbf{\textsc{LIRA} (Ours)}
& 0.5B
& 74.9
& 51.8
& 82.0
& 96.9
& 90.2
& 80.5
& 78.1
& \textbf{78.0} \\

\midrule
\multicolumn{10}{c}{\textit{Fine-tuned on LIBERO-Plus}} \\
\midrule

OpenVLA-OFT~\citep{kim2025fine}
& 7B
& 92.8
& 30.3
& 85.8
& 94.9
& 93.9
& 89.3
& 77.6
& 79.5 \\

VLA-Adapter~\citep{wang2026vlaadapter}
& 0.5B
& 96.7
& 42.1
& 71.1
& 96.5
& 97.5
& 96.9
& 74.4
& 81.0 \\

\textbf{\textsc{LIRA} (Ours)}
& 0.5B
& 93.5
& 51.5
& 83.0
& 95.8
& 91.4
& 92.6
& 77.7
& \textbf{82.9} \\

\bottomrule
\end{tabular*}
}

\caption{
LIBERO-Plus success rates (\%) under zero-shot transfer and
target-domain fine-tuning. Param. is in billions.
Overall is computed over all 10,030 LIBERO-Plus tasks rather than as
the unweighted mean of the seven category-level success rates.
The best Overall result within each setting is shown in bold.
}
\label{tab:libero_plus}
\end{table*}

\section{Experiments}
\label{sec:experiments}

\subsection{Experimental Setup}
\label{subsec:setup}

\paragraph{Benchmarks and metrics.}
We evaluate \textsc{LIRA} on LIBERO~\citep{liu2023libero},
LIBERO-Plus~\citep{fei2026liberoplus}, and CALVIN
ABC$\rightarrow$D~\citep{mees2022calvin}.
LIBERO contains four task suites, while LIBERO-Plus contains 10,030
perturbed tasks spanning seven controlled perturbation categories.
We evaluate LIBERO-Plus under both zero-shot transfer and target-domain
fine-tuning settings.
CALVIN trains on environments A, B, and C and evaluates sequential task
completion in environment D.
We report success rates for LIBERO and LIBERO-Plus; for CALVIN, we also
report the average completed sequence length.

\paragraph{Baselines and controlled comparison.}
Reported VLA systems provide benchmark context across model scales and
training pipelines.
VLA-Adapter~\citep{wang2026vlaadapter} serves as the controlled
architectural baseline, while VLA-Adapter-Pro provides a stronger
compact reference from the same model family.

\paragraph{Implementation details.}
Unless otherwise stated, the compact comparisons use a Prismatic-style
VLM with a Qwen2.5-0.5B language backbone~\citep{yang2024qwen25}, two
image observations, proprioceptive input, LoRA fine-tuning, and
supervised continuous-action regression.
We use LoRA rank 64, a learning rate of $1\times10^{-4}$, a batch size
of 16, and four NVIDIA H100 GPUs.
All main experiments use $M=64$ LIRA Query tokens and a centered
three-layer routing window ($r=1$), except for the corresponding
ablation studies.
For LIBERO-Plus, we follow the official evaluation protocol and perform
one rollout for each of the 10,030 benchmark tasks.
Evaluation otherwise uses 500 trials per LIBERO suite and 1,000 task
sequences on CALVIN.
All ablations follow the same compact training and evaluation protocol
unless stated otherwise.
Complete implementation and evaluation details are provided in the
appendix. Code and model checkpoints will be released upon publication.

\subsection{Main Benchmark Results}
\label{subsec:main_results}

\paragraph{Standard manipulation tasks.}
Table~\ref{tab:libero_main} shows that \textsc{LIRA} improves the direct
VLA-Adapter baseline on all four LIBERO suites.
The average success rate increases from $97.3\%$ to $98.9\%$, with the
largest gain on LIBERO-Long, from $95.0\%$ to $97.6\%$.
Relative to VLA-Adapter-Pro, \textsc{LIRA} raises the average by
$0.4$ percentage points and LIBERO-Long by $1.2$ percentage points.
The largest controlled improvement therefore occurs on the multi-stage
suite, motivating the complementary sequential evaluation on CALVIN.

\paragraph{Sequential task completion.}
Table~\ref{tab:calvin_main} shows that the advantage of \textsc{LIRA}
over VLA-Adapter becomes more pronounced at greater sequence depths.
The five-task success rate increases from $76.5\%$ to $80.5\%$, while
the average completed sequence length rises from $4.42$ to $4.52$.
Compared with VLA-Adapter-Pro, \textsc{LIRA} gains $0.5$ percentage
points on five-task completion and $0.02$ in average sequence length.
Among the compared methods, \textsc{LIRA} achieves the highest success
rates for completing two through five consecutive tasks and the highest
average completed sequence length, while VLA-Adapter obtains the highest
single-task success rate.

\paragraph{Robustness under controlled perturbations.}
Table~\ref{tab:libero_plus} provides the clearest evidence for the
proposed routing interface.
Without LIBERO-Plus fine-tuning, \textsc{LIRA} raises the Overall score
of VLA-Adapter from $59.1\%$ to $78.0\%$, an improvement of
$18.9$ percentage points.
It also exceeds OpenVLA-OFT, the strongest compared zero-shot method,
by $8.4$ percentage points.
Across the seven displayed perturbation categories, \textsc{LIRA}
achieves the highest success rate in six, with Background being the
only exception.

After LIBERO-Plus fine-tuning, \textsc{LIRA} obtains the highest
Overall score of $82.9\%$, compared with $81.0\%$ for VLA-Adapter and
$79.5\%$ for OpenVLA-OFT.
These results support improved robustness under the evaluated
perturbations, but do not imply unrestricted out-of-distribution
generalization.

\begin{figure*}[!t]
\centering
\includegraphics[width=\textwidth]{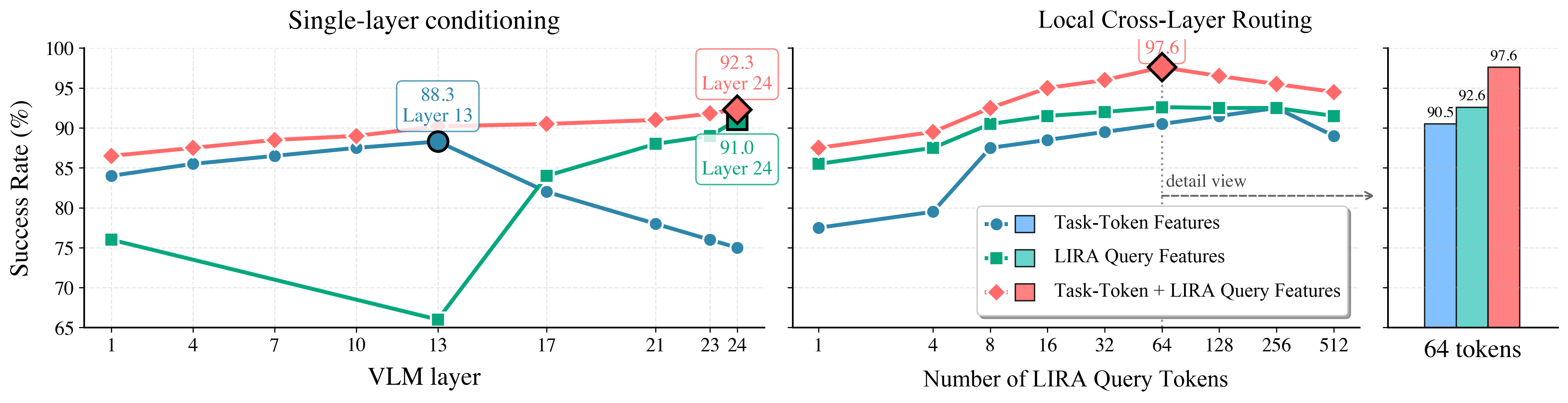}
\caption{
Conditioning-source, VLM-layer, and LIRA Query-token ablations on
LIBERO-Long. Combining task-token features with locally routed LIRA
Query features performs best.
}
\label{fig:ablation_layers_queries}
\end{figure*}

\subsection{Ablation Studies}
\label{subsec:ablation}

\paragraph{Conditioning source and query budget.}
Figure~\ref{fig:ablation_layers_queries} shows that combining
depth-aligned task-token features with locally routed LIRA Query
features outperforms either conditioning source alone.
Table~\ref{tab:query_budget} tests whether this gain can instead be
explained by increasing the number of LIRA Query tokens.

\begin{table}[t]
\centering
\small
\renewcommand{\arraystretch}{1.05}
\begin{tabular*}{\columnwidth}{
    @{\extracolsep{\fill}}p{0.42\columnwidth}cc@{}
}
\toprule
Setting & Query budget & Success rate \\
\midrule
\multirow{8}{=}{Last-layer LIRA Query}
& 1   & 78.0 \\
& 4   & 79.2 \\
& 8   & 87.6 \\
& 16  & 88.7 \\
& 64  & 90.2 \\
& 128 & 91.2 \\
& 256 & 92.8 \\
& 512 & 89.7 \\
\midrule
\multirow{2}{=}{\textsc{LIRA}}
& 64  & \textbf{97.6} \\
& 256 & 93.0 \\
\bottomrule
\end{tabular*}
\caption{
LIRA Query-token budget ablation on LIBERO-Long (\%).
}
\label{tab:query_budget}
\end{table}

The best last-layer-only result reaches $92.8\%$ with 256 tokens,
whereas \textsc{LIRA} reaches $97.6\%$ with only 64 tokens, a gain of
$4.8$ percentage points.
Increasing the \textsc{LIRA} budget to 256 tokens reduces success to
$93.0\%$, indicating that the gain arises from local cross-layer
routing rather than query-token count alone.

\paragraph{Routing topology and window width.}
Tables~\ref{tab:topology_ablation} and~\ref{tab:window_width} compare
the proposed local window with alternative routing topologies and
aggregation widths.

\begin{table}[ht]
\centering
\small
\begin{tabular*}{\columnwidth}{@{\extracolsep{\fill}}lcc@{}}
\toprule
Routing topology & LIBERO-Long & CALVIN-5 \\
\midrule
Flashback routing & 94.5 & 75.7 \\
Column-wise routing & 91.1 & 73.8 \\
Q-Former aggregation & 89.2 & 78.4 \\
\midrule
Local window (\textsc{LIRA}) & \textbf{97.6} & \textbf{80.5} \\
\bottomrule
\end{tabular*}
\caption{
Routing-topology ablation. Success rates are reported in \%.
}
\label{tab:topology_ablation}
\end{table}

\begin{table}[ht]
\centering
\small
\begin{tabular*}{\columnwidth}{@{\extracolsep{\fill}}lcc@{}}
\toprule
Configuration & LIBERO-Long & CALVIN-5 \\
\midrule
Matched layer only & 95.4 & 77.9 \\
Two-layer neighborhood & 96.2 & 79.3 \\
Three-layer neighborhood (\textsc{LIRA})
& \textbf{97.6} & \textbf{80.5} \\
Four-layer neighborhood & 96.4 & 79.8 \\
Global aggregation & 93.7 & 76.9 \\
\bottomrule
\end{tabular*}
\caption{
Routing-window width ablation (\%). Windows are centered on the matched
VLM layer.
}
\label{tab:window_width}
\end{table}

\begin{table}[ht]
\centering
\small
\begin{tabular*}{\columnwidth}{@{\extracolsep{\fill}}lccc@{}}
\toprule
Metric & OpenVLA-OFT & \textsc{LIRA} & Relative \\
\midrule
Backbone size $\downarrow$ & 7B & 0.5B & $1/14\times$ \\
Training memory $\downarrow$ & 62GB & 12.8GB & $0.21\times$ \\
Throughput $\uparrow$ & 71.4Hz & 186.3Hz & $2.61\times$ \\
LIBERO avg. $\uparrow$ & 97.1\% & 98.9\% & +1.8 points \\
\bottomrule
\end{tabular*}
\caption{
System-level resource comparison.
Throughput uses 8-dimensional action chunks, and training memory is
measured with batch size 8.
}
\label{tab:efficiency}
\end{table}

The local window exceeds the strongest alternative by $3.1$ points on
LIBERO-Long and $2.1$ points on CALVIN-5.
The centered three-layer window performs best on both benchmarks;
expanding it to four layers provides no benefit, while global
aggregation performs worst.
These results favor locally structured routing over unrestricted
cross-layer aggregation.

\begin{table*}[t]
\centering

{\small
\setlength{\tabcolsep}{6pt}
\begin{tabular*}{\textwidth}{
    @{\extracolsep{\fill}}lcccccc@{}
}
\toprule
\multirow{2}{*}{Method}
& \multirow{2}{*}{Move}
& \multirow{2}{*}{Pick-and-Place}
& \multicolumn{2}{c}{Collection}
& \multicolumn{2}{c}{Transfer} \\
\cmidrule(lr){4-5}
\cmidrule(lr){6-7}
& & & Step 1 & Step 2 & Step 1 & Step 2 \\
\midrule

OpenVLA-OFT~\citep{kim2025fine}
& 8/10 & 6/10 & 8/10 & 5/10 & 6/10 & 4/10 \\

VLA-Adapter~\citep{wang2026vlaadapter}
& 9/10 & 7/10 & 9/10 & 6/10 & \textbf{8/10} & 6/10 \\

\textbf{\textsc{LIRA} (Ours)}
& \textbf{10/10}
& \textbf{9/10}
& \textbf{10/10}
& \textbf{8/10}
& \textbf{8/10}
& \textbf{7/10} \\

\bottomrule
\end{tabular*}
}

\caption{
Real-world success over 10 trials per task.
For two-stage tasks, Step 1 reports intermediate success and Step 2
reports final task success.
}
\label{tab:realworld}
\end{table*}

\subsection{Resource Profile}
\label{subsec:efficiency}

Table~\ref{tab:efficiency} compares the practical resource profiles of
\textsc{LIRA} and OpenVLA-OFT.
Because the two systems use different backbones and training protocols,
we treat this as a system-level comparison of their reported
configurations.

Under the reported configurations, \textsc{LIRA} uses a
14$\times$ smaller backbone, requires 79\% less training memory, and
achieves 2.61$\times$ higher inference throughput than OpenVLA-OFT,
while improving the LIBERO average by 1.8 percentage points.
These results show that the evaluated \textsc{LIRA} configuration
combines strong manipulation performance with a compact resource
profile.


\begin{figure}[!htbp]
\centering
\includegraphics[width=\linewidth]{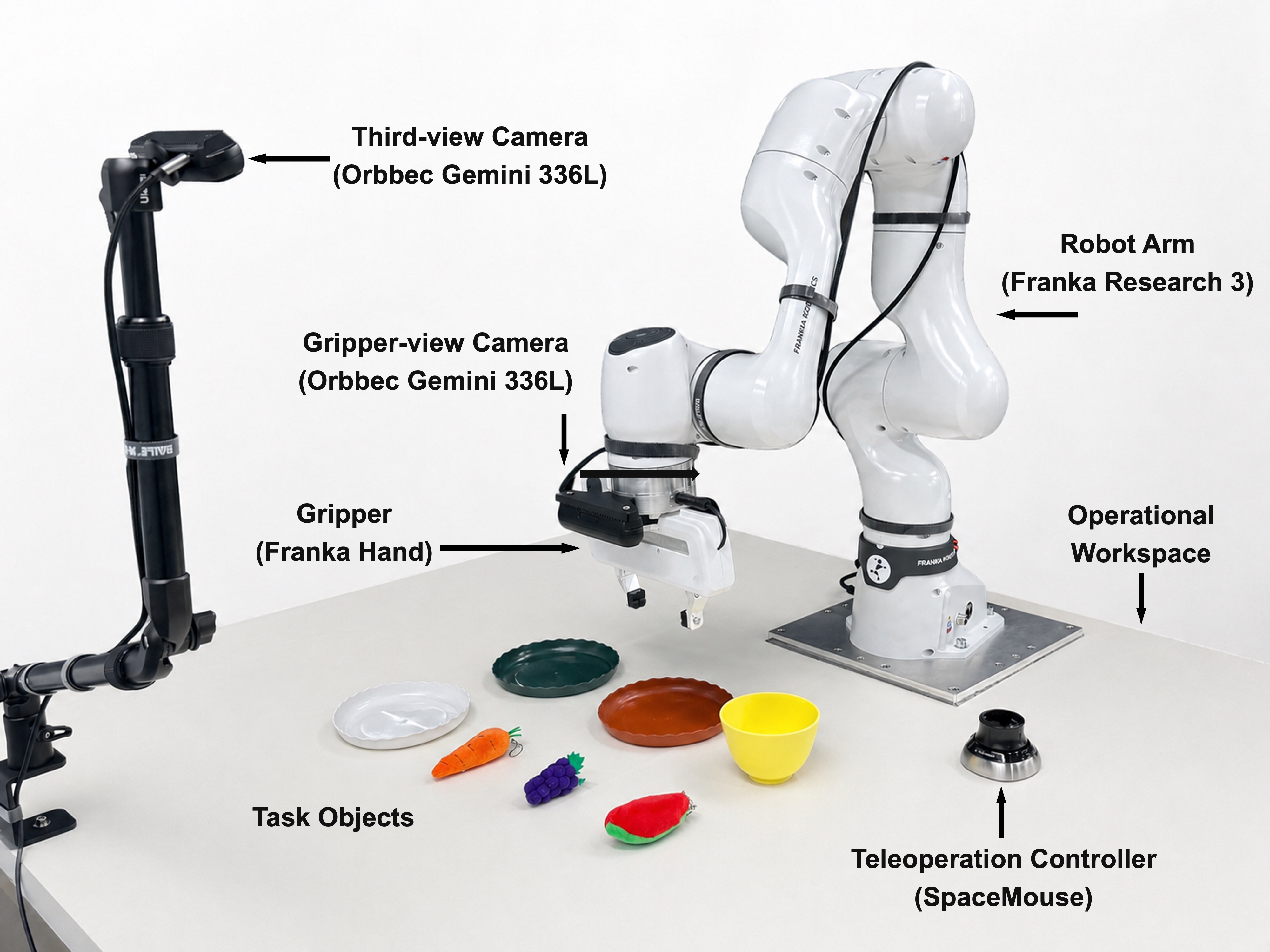}
\caption{
Real-world experimental platform with a Franka Research 3 arm,
third-person and gripper-view cameras, and a SpaceMouse teleoperation
interface.
}
\label{fig:realworld_platform}
\end{figure}

\subsection{Real-World Evaluation}
\label{subsec:realworld}

We further evaluate \textsc{LIRA} on a physical robot platform using
spatial, pick-and-place, and two-stage manipulation tasks.
This evaluation tests whether the benefits of local cross-layer
routing extend beyond the simulation benchmarks.

\paragraph{Platform and demonstrations.}
As shown in Figure~\ref{fig:realworld_platform}, the platform consists
of a 7-DoF Franka Research 3 arm with a Franka Hand gripper.
Two Orbbec Gemini 336L cameras provide third-person and gripper-view
images.
Human demonstrations are collected using a SpaceMouse teleoperation
interface.

\paragraph{Tasks and success criteria.}
We evaluate four tasks of increasing compositional complexity.
\textit{Move} requires moving the gripper to the left or right of a
target object and is considered successful when the final gripper
position satisfies the instructed spatial relation.
\textit{Pick-and-Place} requires grasping a specified object and
placing it into a target plate; success requires the correct object to
be released inside the plate.
\textit{Collection} is a two-stage task in which the robot first places
a plate on the instructed side of the table and then places a specified
object into that plate.
\textit{Transfer} requires first placing one specified object into the
plate and then placing a second specified object into the same plate.
For the two-stage tasks, Step 1 denotes completion of the intermediate
subgoal, whereas Step 2 denotes completion of the full instruction.

We compare \textsc{LIRA} with OpenVLA-OFT~\citep{kim2025fine} and
VLA-Adapter~\citep{wang2026vlaadapter} under the same task definitions.
Table~\ref{tab:realworld} reports success over 10 trials per task.

\paragraph{Results.}
As reported in Table~\ref{tab:realworld}, \textsc{LIRA} achieves the
highest final success on all four tasks, including the two-stage
instructions that require carrying information across subgoals.
Compared with VLA-Adapter, it improves Pick-and-Place success from
7/10 to 9/10, Collection Step 2 from 6/10 to 8/10, and Transfer Step 2
from 6/10 to 7/10.
These results show that the benefit of local cross-layer routing
extends to the evaluated physical-robot setting.
Given 10 trials per task, these results should be regarded as
preliminary.

\section{Conclusion}
\label{sec:conclusion}

We presented \textsc{LIRA}, a local cross-layer action-conditioning
mechanism for vision-language-action models.
By formulating VLM-to-action conditioning as depth-aware information
routing, \textsc{LIRA} introduces a locality prior over VLM depth.
For each PFB, it routes LIRA Query features from a depth-aligned local
window while retaining the layer-aligned task-token branch.
This local cross-layer routing introduces no additional trainable
parameters relative to VLA-Adapter.

Under a matched 0.5B-backbone configuration, \textsc{LIRA} improves the
VLA-Adapter baseline across LIBERO, CALVIN ABC$\rightarrow$D, and
LIBERO-Plus.
In zero-shot transfer to LIBERO-Plus, it raises overall task-level
success from $59.1\%$ to $78.0\%$.
Ablations on routing topology, routing-window width, and LIRA Query-token
budget further support the centered three-layer window over
one-to-one layer-aligned LIRA Query routing, alternative routing
topologies, and global aggregation.
The real-world study provides complementary evidence beyond simulation.

The current evidence remains limited to one compact VLM backbone, one
PFB-based action-decoder architecture, the evaluated LIBERO and CALVIN
protocols, and a single-robot study with 10 trials per task.
It therefore does not establish cross-embodiment robustness or
performance on contact-rich and force-sensitive manipulation.
Future work should evaluate local cross-layer routing across different
VLM backbones, robot embodiments, and more complex manipulation
settings.

\newpage
\bibliography{main}


\end{document}